\documentclass[conference]{IEEEtran}
\IEEEoverridecommandlockouts
\usepackage{cite}
\usepackage{flushend}
\usepackage{amsmath,amssymb,amsfonts}
\usepackage{algorithmicx}
\usepackage{graphicx}
\usepackage{textcomp}
\usepackage{xcolor}
\usepackage[noend]{algpseudocode}      
\usepackage{mathtools}
\usepackage{balance}
\usepackage{amsmath}
\def\BibTeX{{\rm B\kern-.05em{\sc i\kern-.025em b}\kern-.08em
    T\kern-.1667em\lower.7ex\hbox{E}\kern-.125emX}}

\usepackage[caption=false,labelformat=empty]{subfig}
\usepackage{color}
\usepackage{listings}
\usepackage{caption}

\newcounter{nalg} 
\renewcommand{\thenalg}{\arabic{nalg}} 
\DeclareCaptionLabelFormat{algocaption}{Algorithm \thenalg} 

\lstnewenvironment{algorithm}[1][] 
{   
    \refstepcounter{nalg} 
    \captionsetup{labelformat=algocaption,labelsep=colon} 
    \lstset{ 
        mathescape=true,
        frame=tB,
        numbers=left, 
        numberstyle=\tiny,
        basicstyle=\scriptsize, 
        keywordstyle=\color{black}\bfseries\em,
        keywords={,input, output, return, datatype, function, in, if, else, for, each, while, begin, end, } 
        numbers=left,
        xleftmargin=.04\textwidth,
        breaklines=true,
        #1 
    }
}
{}

\begin{document}

\title{Approximate kNN Classification \\for Biomedical Data\\
\thanks{This project has received funding from the Hellenic Foundation for Research and Innovation (HFRI) and the General Secretariat for Research and Technology (GSRT), under grant agreement No 1901.}
}

\makeatletter
\newcommand{\linebreakand}{
  \end{@IEEEauthorhalign}
  \hfill\mbox{}\par
  \mbox{}\hfill\begin{@IEEEauthorhalign}
}
\makeatother

\author{\IEEEauthorblockN{Panagiotis Anagnostou}
\IEEEauthorblockA{\textit{Dept. of Computer Science} \\
\textit \& \textit{Biomedical Informatics} \\
\textit{University of Thessaly}\\
Lamia, Greece \\
panagno@uth.gr} 
\and
\IEEEauthorblockN{Petros Barbas}
\IEEEauthorblockA{\textit{Dept. of Computer Science} \\
\textit \& \textit{Biomedical Informatics} \\
\textit{University of Thessaly}\\
Lamia, Greece \\
 petrosbarmpas@uth.gr}
\linebreakand
\IEEEauthorblockN{Aristidis G. Vrahatis}
\IEEEauthorblockA{\textit{Dept. of Computer Science} \\
\textit \& \textit{Biomedical Informatics} \\
\textit{University of Thessaly}\\
Lamia, Greece \\
arisvrahatis@uth.gr}
\and
\IEEEauthorblockN{Sotiris K. Tasoulis}
\IEEEauthorblockA{\textit{Dept. of Computer Science} \\
\textit \& \textit{Biomedical Informatics} \\
\textit{University of Thessaly}\\
Lamia, Greece \\
stasoulis@uth.gr}

}

\maketitle

\begin{abstract}
We are in the era where the Big Data analytics has changed the way of interpreting the various biomedical phenomena, and as the generated data increase, the need for new machine learning methods to handle this evolution grows. An indicative example is the single-cell RNA-seq (scRNA-seq), an emerging DNA sequencing technology with promising capabilities but significant computational challenges due to the large-scaled generated data. Regarding the classification process for scRNA-seq data, an appropriate method is the k Nearest Neighbor (kNN) classifier since it is usually utilized for large-scale prediction tasks due to its simplicity, minimal parameterization, and model-free nature. However, the ultra-high dimensionality that characterizes scRNA-seq impose a computational bottleneck, while prediction power can be affected by the "Curse of Dimensionality". In this work, we proposed the utilization of approximate nearest neighbor search algorithms for the task of kNN classification in scRNA-seq data focusing on a particular methodology tailored for high dimensional data. We argue that even relaxed approximate solutions will not affect the prediction performance significantly. The experimental results confirm the original assumption by offering the potential for broader applicability.

\end{abstract}

\section{Introduction}

The field of Biomedicine is in the era of data explosion from a plethora of emerging technologies such as the high throughput sequencing, which has revolutionized the field of molecular biology \cite{slatko2018overview}. The upcoming technology of this category is considered the optimized Next Generation Sequencing (NGS) called single-cell technologies \cite{eberwine2014promise}. It studies the sequence information from individual cells, providing a higher resolution of cellular differences and a better comprehension of the function of individual cells in their micro-environment \cite{kolodziejczyk2015technology}.

Regarding the transcriptomics area, single-cell RNA-sequencing (scRNA-seq) is a rapidly growing and widely applying technology with great potential \cite{reuter2015high}. From each cell, we obtain expression measurements for each gene; thus, it offers the opportunity to uncover the complexity of tissues' cellular heterogeneity \cite{chen2018tissues,buettner2015computational,li2017reference}. However, this potential presupposes the proper scRNA-seq data analysis, which is quite challenging since we now have experimental studies with an enormous volume in both cell samples and gene expression profiles. 
Hence, despite the popularity and the impact of scRNA-seq, analyzing such data remains a challenging task. Besides their volume and complexity, such data have an inherent noise in gene expression measurements caused by the low capture efficiency \cite{stegle2015computational}. Also, several zero counts are obtained from scRNA-seq experiments since it often relies on low expression counts. This limitation creates the dropout effect by adding more complexity to their computational analysis.

A challenging part of scRNA-seq data analysis is the classification process under the machine learning perspective since we have the potential to identify the various cell populations in a more comprehensive way \cite{svensson2018exponential, tasoulis2018biomedical}. This process belongs to the cell level challenges \cite{vrahatis2020recent}, having the potential to uncover the inherent cellular heterogeneity by offering a promising de novo discovery \cite{hedlund2018single}. Towards this direction, several classification algorithms tailored for single-cell RNA-seq data have been recently proposed  \cite{alquicira2019scpred, qi2020clustering, vrahatis2020ensemble}. These tools have promising results, however, their limitation is that they ignore the ultra-high volume in both samples (cells) and dimensions (genes) \cite{angerer2017single}. It is worth mentioning that at the beginning of the single-cell technology in previous years, the majority of such experiments produced a few tens or hundreds of cell samples with tens of thousand genes measurements. In their analysis, these were considered small n large - p problems, where n and p are the cells and genes, respectively. Now, the single-cell technological evolution has shifted their analysis into large n - large p problems. There are now experimental studies with several thousands of cell samples, some even reaching 1 million cells, such as the 10x Genomics biotechnology company. 

It is clear that scRNA-seq data has met the big data family, having an ultra-high volume in both samples and dimensions size. Two leading solutions prevail for handling this complexity, high-performance computing (HPC), and the approximate search frameworks. HPC has remarkable progress in the entire next-generation sequencing domain with a large variety of published tailored methods \cite{schmidt2017next}. As these solutions require specialized infrastructure and huge computational resources, it is necessary to show exceptional interest in the approximate searching solution.

In this work, we study the operation of cutting-edge approximate nearest neighbor search methods in single-cell RNA-seq data when incorporated into the kNN classification method, seeking to expose their usability and reliability in managing the complexity of scRNA-seq data. We emphasize on analyzing the behavior of the MRPT \cite{hyvonen2016fast}, a recent approximate search algorithm tailored for high dimensional data, with respect to its effect  on classification accuracy.

\section{Approximate k-Nearest Neighbor search}
k-Nearest Neighbor searching is still a very challenging task for high-dimensional data. In an effort to balance the efficiency and the accuracy, a number of c-approximate nearest neighbor (c-ANN) algorithms have been introduced to return an approximation of Nearest Neighbors with confidence of at least $\delta$. To speed up the search these methods are preprocessing the data into an efficient index.
Initially, before indexing usually the data are compressed through a dimensionality reduction method such as Principal Component Analysis \cite{buettner2014probabilistic} ,Random Projection spaces \cite{xie2016comparison} or Feature Selection \cite{townes2019feature} in an attempt to uncover their “Intrinsic” dimensionality. Then the actual index is constructed using data structure-based techniques like Trees, Hashing, and Quantization that manage to encode vectors in a much more compact form. Finally, given a query point, the non-exhaustive search is applied to retrieve an approximation of its nearest neighbors.

\paragraph{Hashing} Hashing based algorithms transform every data point to a low-dimensional representation, so each point can be represented by a shortcode called hash code. There are two main hashing sub-categories: Locality Sensitive Hashing (LSH) \cite{pan2011fast} and Learning to Hash (L2H) \cite{cao2019general}.
Locality sensitive hashing is a data-independent hashing approach. The LSH methods rely on a family of locality-sensitive hash functions that map similar input data points to the same hash codes with higher probability than different points. Random linear projections are commonly used by hash functions to generate the hash code.

Learning to Hash methods use the data distribution to generate specific hash functions, leading to higher efficiency at the cost of giving away the theoretical guarantees.

\paragraph{Partition Based} Methods in this category that encapsulates Trees can be seen as dividing the entire high dimensional space into multiple disjoint subspaces. If a query "q" is located in a region Rq, then its nearest neighbors should belong in Rq or regions close to Rq. The partition process carries out recursively, building a tree structure. There are types of partition based methods: pivoting, hyperplane, and compact partitioning schemes. Pivoting methods divide the points relying on the distances from the points to some pivots. A representative example is the Ball Tree algorithm \cite{dolatshah2015ball}. Hyperplane partitioning methods recursively divide the space by a hyperplane usually defined by a random direction. Representative examples of this category are the Annoy \cite{Bernhardsson2015}, the Random-Projection Forest \cite{dasgupta2008random} and the Randomized KD-trees  \cite{silpa2008optimised} algorithms. Compact partitioning algorithms either divide the data into clusters or create approximate Voronoi partitions \cite{beygelzimer2006cover} to exploit locality.

\paragraph{Graph Based} Graph-based methods construct a proximity graph where each data point corresponds to a node, while edges connecting some of these nodes define the neighbor relationship. The core concept is that a neighbor’s neighbor is also likely to be a neighbor. The search can be efficiently performed by iteratively expanding neighbors’ neighbors in a best-first search strategy following the edges. Differences between the graph structures define different variations of Graph-based methods.

\begin{figure*}[t]
	\centering
	\subfloat[][(a)]{
		\includegraphics[width=2.3in]{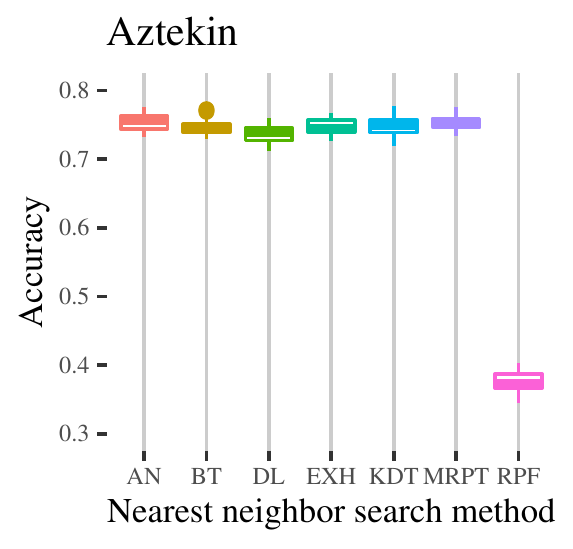}
		\label{fig:comp_acc_a}
	}%
	\subfloat[][(b)]{
		\includegraphics[width=2.3in]{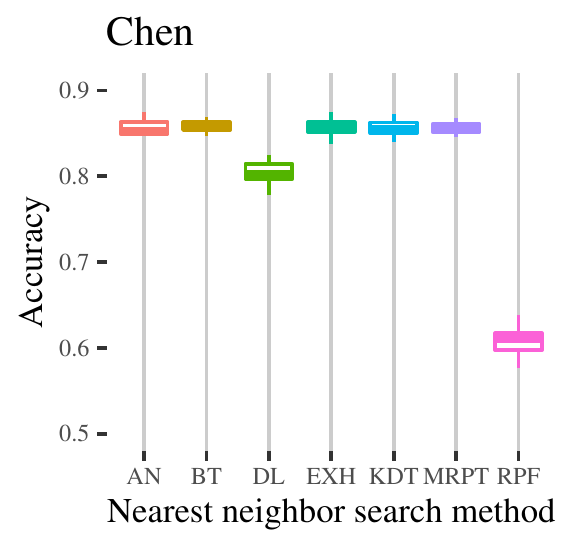}
		\label{fig:comp_acc_b}
	}%
	\subfloat[][(c)]{
		\includegraphics[width=2.3in]{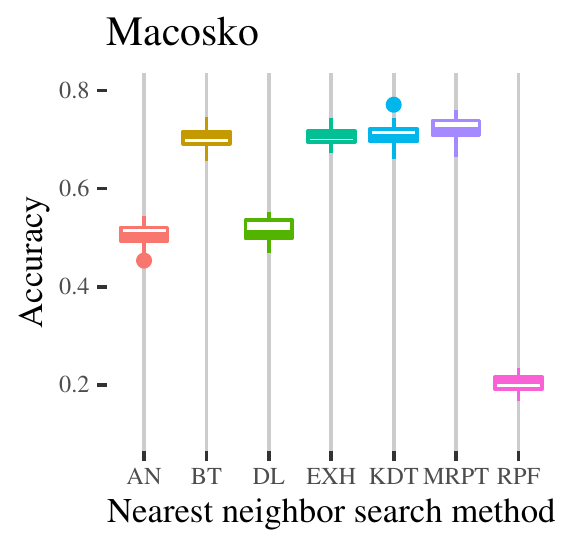}
		\label{fig:comp_acc_c}
	}
	\caption{Boxplots depict the effect of various approximate nearest neighbor search methods on the kNN classification algorithm in three publicly available single-cell RNA-seq data (Aztekin (a), Chen (b), and Macosko (c)). The classification performance is evaluated in terms of accuracy utilizing six other methods besides the MRPT framework. Four approximate nearest neighbor search methods (Annoy (AN), Ball Tree (BT), DolphinPy (DL), and RP Forrest (RPF)) were utilized along with the traditional exhaustive search (EXH) and the well-known kd-tree (KDT) method. All methods were applied on 100 independent experiments using the 10-fold cross-validation technique \label{fig:comp_acc}}
\end{figure*}

\begin{figure*}[t]
	\centering
	\subfloat[][(a)]{
		\includegraphics[width=2.3in]{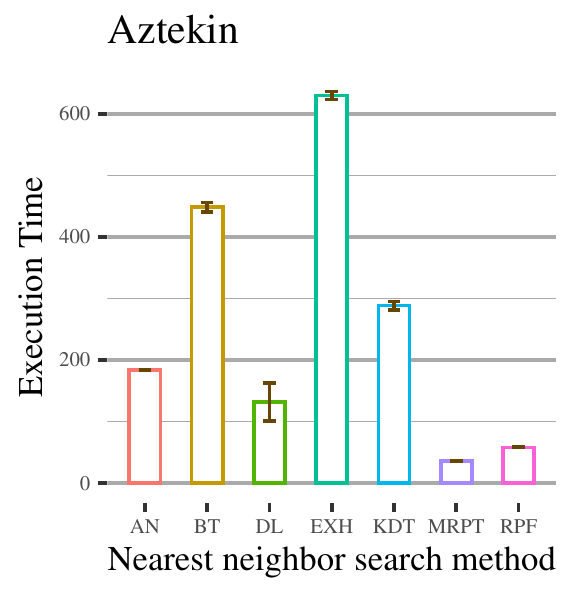}
		\label{fig:comp_time_a}
	}%
	\subfloat[][(b)]{
		\includegraphics[width=2.3in]{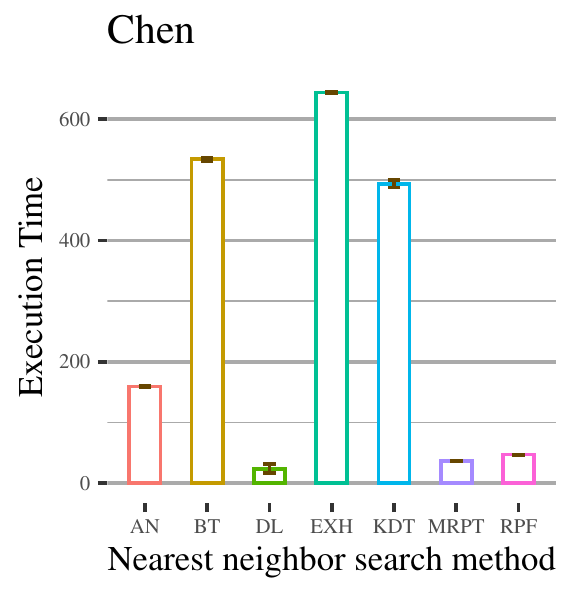}
		\label{fig:comp_time_b}
	}%
	\subfloat[][(c)]{
		\includegraphics[width=2.3in]{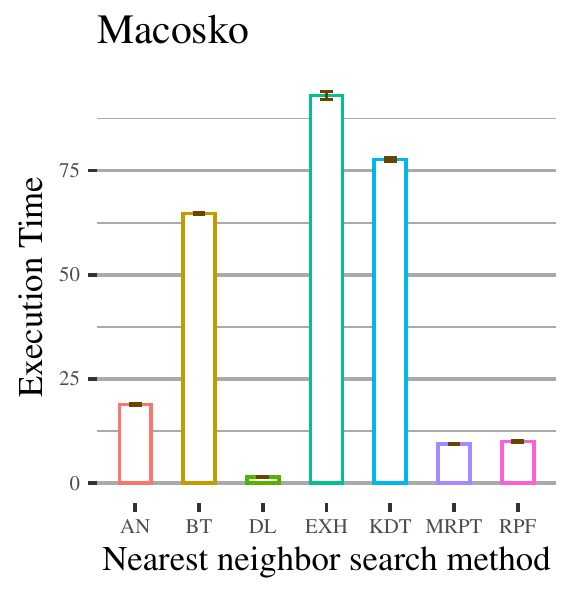}
		\label{fig:comp_time_c}
	}
	\caption{Error bars imprint the execution time of all seven kNN classification schemes using the comparative nearest neighbor search methods (Annoy (AN), Ball Tree (BT), DolphinPy (DL), kd-tree (KDT), Exhaustive (EXH), MRPT, and RP Forrest (RPF)), in the three publicly available datasets Aztekin (a), Chen (b), and Macosko (c). All methods were applied on 100 independent experiments using the 10-fold cross-validation technique.
\label{fig:comp_time}}
\end{figure*}

\section{Speeding up the kNN classification}

kNN classification along with its multiple extensions/variations is still widely used in several real world applications due to its simplicity and effectiveness. Its most critical advantage is that there is no need to build a model, tune several parameters, or make additional assumptions. Additionally, the algorithm is versatile, thus it can be used for both classification and regression tasks.
However, as the volume of data increases the algorithm becomes significantly slower becoming an impractical choice in environments where predictions need to be made rapidly. For this reason in this work we employ approximate nearest neighbor search for speeding up the kNN classification procedure. We particularly focus on the MRPT algorithm \cite{hyvonen2016fast} that has been proven to be the faster approximate method for very high dimensional data. 
We argue that the approximate nearest neighbor search integrated within a kNN classifier will not affect its classification performance negatively while the MRPT algorithm's characteristics will allow us to tackle the high complexity and dimensionality that characterises the single cell data.

MRPT starts with creating a binary space-partitioning tree index, utilizing the auto-tuning process proposed in \cite{jaasaari2019efficient}. This process utilizes the sparse random-projections trees (sRPT) for which the parametrization is done according to the nearest neighbors' target recall level, provided to the algorithm along with the ($k$) number of nearest neighbors to search for. 

The search within the index for a query point $q$ is achieved by assigning the query point to a leaf for all constructed trees. For this purpose at each node of the tree the query point is projected onto a random direction and subsequently, a decision is taken concerning whether it should be assigned to the left or right child node based on a splitting criterion.
The information retrieved from each leaf is combined through a voting process to provide the final estimation.
Note here that the method can be considered almost parameter-free since the number of trees $T$, the depth of the trees $l$, and the vote threshold $v$ accordingly are automatically set given the desired recall rate.

It is worth mentioning that the target recall defines the degree of confidence in retrieving the actual nearest neighbors through an exhaustive search. A recall rate equal to 1 will force the algorithm to perform an exhaustive nearest neighbor search.

\begin{table*}[t]
\centering
\caption{Mean values of the classification performance evaluation techniques for the three datasets Aztekin, Chen, and Macosko, of all seven kNN classification schemes using the comparative nearest neighbor search methods (Annoy (AN), Ball Tree (BT), DolphinPy (DL), kd-tree (KDT), Exhaustive (EXH), MRPT, and RP Forrest (RPF)). All methods were applied on 100 independent experiments using the 10-fold cross-validation technique.}
\label{tbl:accunulative_results}
\begin{tabular}{l|ccccccc}
\hline
     & Accuracy & Sencitivity & Specificity & Precision & Recall & F1-score & Execution Time  \\
\hline
Annoy   & 0.7058            & 0.5856      & 0.9730      & 0.7062    & 0.5856 & 0.6801   & 120.48          \\
Ball Tree   & 0.7703            & 0.6196      & 0.9738      & 0.7961    & 0.6196 & 0.7199   & 349.08          \\
DolphinPy   & 0.6850            & 0.4743      & 0.9707      & 0.6155    & 0.4743 & 0.5953   & 52.270           \\
KD-Tree  & 0.7701           & 0.6208      & 0.9738      & 0.7954    & 0.6208 & 0.7229   & 286.60          \\
Exhaustive  & 0.7708           & 0.6201      & 0.9738      & \textbf{0.7979}    & 0.6201 & 0.7168   & 455.70          \\
MRPT & \textbf{0.7770}   & \textbf{0.6252}      & \textbf{0.9745}      & 0.7970    & \textbf{0.6252} & \textbf{0.7241}   & \textbf{27.040}           \\
RP Forrest  & 0.3967           & 0.2511      & 0.9363      & 0.2998    & 0.2511 & 0.3409   & 38.350           \\
\hline
\end{tabular}
\end{table*}

\section{Experimental Results and Discussion}

To examine the approximate nearest neighbor methods' contribution to the single-cell RNA-seq classification process, along with the suggested MRPT \cite{hyvonen2016fast} method, we additionally employ four cutting-edge nearest neighbor search methods. These are the  RP Forrest \cite{Serko2019}, ANNOY \cite{Bernhardsson2015}, DolphinPy \cite{Psaros2017}, and Ball Tree \cite{dolatshah2015ball}. For consistency, they all belong to the family of partition-based approximation methods similarly to MRPT. For reference, these were compared against the original kNN exhaustive search and the well-established KD-Tree search method.

\begin{figure*}[t]
	\centering
	\subfloat[][(a)]{
		\includegraphics[width=2.3in]{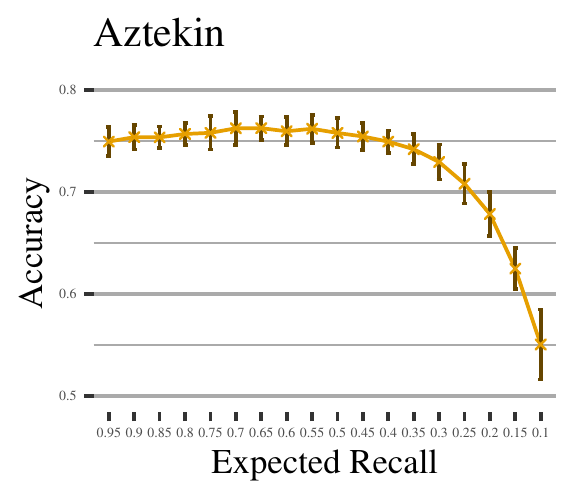}
		\label{fig:mrpt_acc_a}
	}
	\subfloat[][(b)]{
		\includegraphics[width=2.3in]{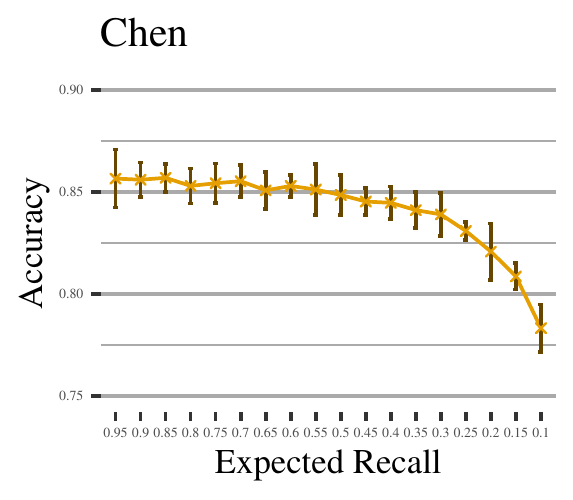}
		\label{fig:mrpt_acc_b}
	}
	\subfloat[][(c)]{
		\includegraphics[width=2.3in]{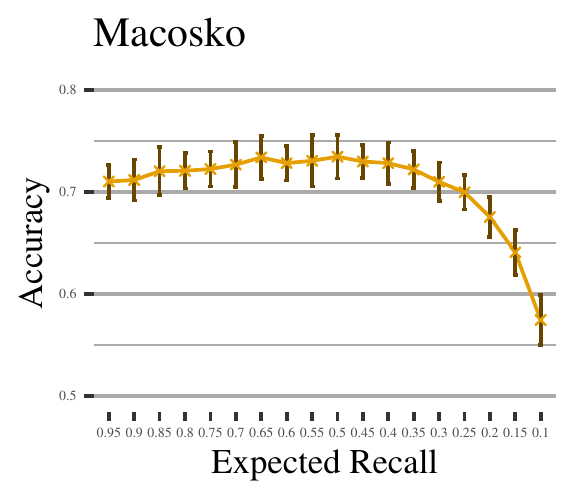}
		\label{fig:mrpt_acc_c}
	}
	\caption{Line plot with error bars depict the effect of the recall rate parameter of MRPT approximate nearest neighbor search method on the kNN classification algorithm. It defines the degree of confidence in retrieving the actual nearest neighbors through an exhaustive search (recall rate equal to 1 will force the algorithm to perform exhaustive nearest neighbor search).}
\end{figure*}

\begin{figure*}[t]
	\centering
	\subfloat[][(a)]{
		\includegraphics[width=2.3in]{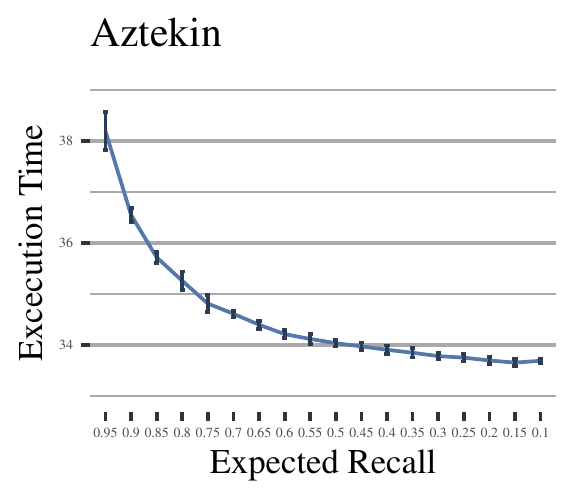}
		\label{fig:mrpt_time_a}
	}
	\subfloat[][(b)]{
		\includegraphics[width=2.3in]{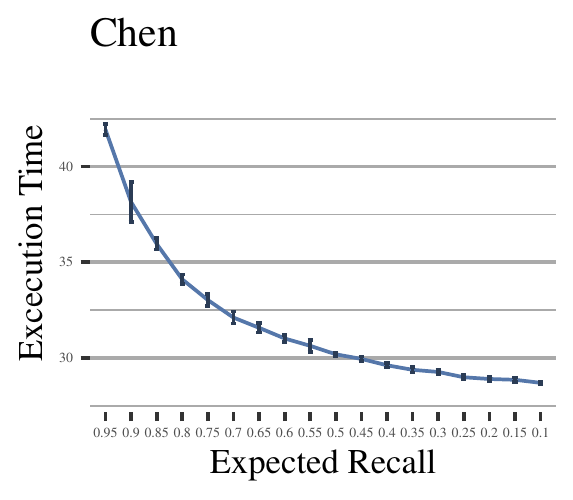}
		\label{fig:mrpt_time_b}
	}
	\subfloat[][(c)]{
		\includegraphics[width=2.3in]{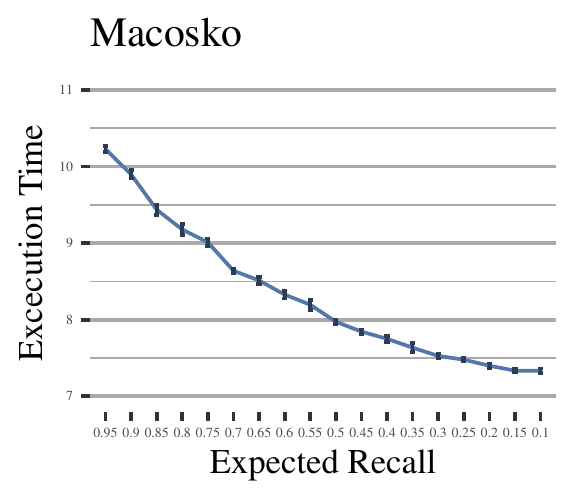}
		\label{fig:mrpt_time_c}
	}
	\caption{Line plot with error bars depict the execution time of MRPT approximate nearest neighbor search method on the kNN classification algorithm with respect to its recall rate parameter. \label{fig:mrpt_time}}
\end{figure*}

More specifically, Ball Tree \cite{dolatshah2015ball,nielsen2009tailored} is based on the idea of metric trees \cite{uhlmann1991metric}, a concept in which data are organized and structured considering the metric space in which they are located, and samples do not need to be of finite dimensions or in vectors \cite{kumar2008good}. The algorithm splits the dataset points into two clusters. Each cluster is enclosed in a hypersphere. The clusters are picked in such a way that they have maximum distance between them. The clusters are allowed to be unbalanced, and the process of separating the data points into two spheres is repeated within every node until a selected depth is reached. This strategy leads to a hierarchy of clusters containing more and more circles.
RP Forrest \cite{Serko2019} creates a forest of k binary random projection trees while for every tree, the training set recursively partitioned into smaller subsets until a leaf node of at most $N$ points is created. Every split is based on the cosine of the angle the points make with a randomly drawn hyperplane. The resulting tree has a max leaf size of $N$ and is almost balanced due to the median splits, resulting in consistent traversing times. Every query response is accomplished by traversing all trees to the query point's leaf node. Then ANN candidates from that tree are retrieved while being merged and sorted by the distance to the query point. This technique differs from other ANN algorithms in that it is not necessary to store all the vectors indexed in the model.
ANNOY \cite{Bernhardsson2015} is a novel technique, which utilizes random projections to construct a tree. In every tree node, the space is binary divided into two sub-spaces by a hyperplane. The hyperplane is chosen by randomly sampling two points from the subset and taking the hyperplane at equal distances from each. The procedure is replicated k times so that a forest is created. Lastly, optimizations, to increase performance, including Hamming distance encoding and Dot product space embedding, are implemented.
Similarly, DolphinPy \cite{Psaros2017} is an efficient method for computing approximate nearest neighbors in higher dimensions. Basically, given a set of locality-sensitive hashing family functions for some metric space, points were randomly projected to the Hamming cube of dimension $log(n)$, where $n$  is the number of input points. The projected space contains strings that serve as keys for buckets containing the input points. The query algorithm projects the query point and then assesses points which are assigned to the same or nearby vertices on the Hamming cube. This method resembles the multi-probe LSH approach, but it differs in how the candidates' list is computed.

All aforementioned tools were applied to three publicly available single-cell RNA sequencing datasets from various transcriptomics evaluation studies. All datasets (henceforth they will be referred to as Aztekin, Chain, Macosko, which are the first author names of the original studies) were obtained from the "scRNAseq" Bioconductor package \cite{risso2020Packages}. The $Aztekin$ data \cite{aztekin2019identification} is the largest database in our analysis which contains $31,535$ gene measurements for $13,199$ cell samples of Xenopus tail organism. The samples are pre-annotated into three different classes. The Chen \cite{chen2017single} dataset is composed of single-cell RNA sequencing gene measurements of adult mice's hypothalamus cells. It includes $14,437$  samples  with  $23,284$  genes. The samples are classified into 47 different classes for neural and non-neural cells. The Macosko \cite{macosko2015highly} dataset consists of $6,418$ samples with $12,822$ genes. The samples are classified into 39 different classes.

Prediction performance was assessed by measuring six characteristic classification measures (accuracy, sensitivity, specificity, precision, recall, and F1-score) to extract the strengths and weaknesses of each method for all three datasets (see Table \ref{tbl:accunulative_results} where results are presented accumulatively). All executions were made using the 10-fold cross-validation process in 100 iterations. The number of nearest neighbors to search ($k$) was set for all the algorithms to the value $5$. All algorithms were run with the corresponding default parameters. The recall rate for MRPT been arbitrary set to $0.85$ across all experiments to guarantee a significant performance boost. Minor variation do not alter the outcome significantly and for completeness a parameter analysis is following later on.
For brevity, only the accuracy score is being reported with respect to classification performance in Figure \ref{fig:comp_acc} along with the corresponding computational execution time in Figure \ref{fig:comp_time} for each individual dataset. We observe the consistent behaviour of MRPT achieving a comparable classification performance against the exhaustive search for all datasets similarly to the Ball Tree (BT) method. RP Forrest (RPF) performs significantly worse while both ANNOY (AN) and DolphinPy (DL) significantly reduce the classification performance of kNN for at least one case. MRPT's robustness is also confirmed with respect to execution time as shown in Figure \ref{fig:comp_time} where we observe significant speed boost compared to the exhaustive search but also the Ball Tree (BT).

The accumulative result presented in Table \ref{tbl:accunulative_results} complies with the aforementioned findings. MRPT achieve best performance across almost all classification metrics while we simultaneously achieving to reduce the computation cost by almost 17 times compared to the exhaustive search.

\paragraph{Parameter Analysis}
To this end we study the effect of the recall rate parameter provided as input to the MRPT algorithm with respect to the classification performance of kNN and the corresponding computational cost. Figure \ref{fig:mrpt_acc_c} illustrates the relation between recall rate and classification accuracy. We interestingly observe that accuracy is retained to that of the exhaustive search for a wide range of values while we even identify a minor increase for recall rates close to $0.6$ for the Aztekin and Macosko datasets. Simultaneously, the significant computational benefits for the range $0.95$ to $0.6$ indicate the method's true potential.

\section{CONCLUSIONS}
In this work, we cope with the big data perspective of biomedical data characterized by ultra-high dimensionality. We focus on the emerging single-cell RNA-seq technology, which, while revolutionizing the interpretation of molecular biology, its data analysis involves several computational challenges due to the large-scale generated datasets in each experimental studies. Such studies produce tens of thousands of cell samples with tens of thousands of features (genes measurements). We focus on the classification perspective of these data by studying the utilization of approximate nearest neighbor search algorithms for kNN classification in scRNA-seq data. The corpus of our analysis is a particular methodology tailored for high dimensional data, called MRPT. We provided evidence that approximate solutions in the nearest neighbor searching process of the kNN classifier offer the opportunity not only for fast and scalable classification frameworks, but also for reliable and robust results.

\section*{ACKNOWLEDGMENT}
This project has received funding from the Hellenic Foundation for Research and Innovation (HFRI), under grant agreement No 1901.

\addtolength{\textheight}{-0cm}  
\bibliographystyle{ieeetr}
\balance
\bibliography{root}

\end{document}